\documentclass[journal]{IEEEtran}
\usepackage{times}
\usepackage{url}
\usepackage[utf8]{inputenc}
\usepackage[small]{caption}
\usepackage{graphicx}
\usepackage{amsmath}
\usepackage{amsthm}
\usepackage{booktabs}
\usepackage{algorithm}
\usepackage{algorithmic}
\usepackage{amssymb}

\usepackage[colorlinks,            linkcolor=green,  citecolor=blue]{hyperref}

\begin{document}
\title{ProxyMix: Proxy-based Mixup Training with Label Refinery for Source-Free Domain Adaptation}

\author{Yuhe Ding,
        Lijun Sheng,
        Jian Liang$^\dagger$,
        Aihua Zheng
        and~Ran~He,~\IEEEmembership{Senior~member,~IEEE}
\thanks{$^\dagger$ Jian Liang is the corresponding author.}
\thanks{Yuhe Ding is with the Anhui Provincial Key Laboratory of Multimodal Cognitive Computation, Anhui University and the Center for Research on Intelligent Perception and Computing, Institute of Automation, Chinese Academy of Sciences (CASIA). E-mail: madao3c@foxmail.com.}
\thanks{Lijun Sheng is 
with the University of Science and Technology of China and the Center for Research on Intelligent Perception and Computing, CASIA. E-mail: slj0728@mail.ustc.edu.cn.}
\thanks{Aihua Zheng is with the Anhui Provincial Key Laboratory of Multimodal Cognitive Computation, Anhui University, China. E-mail: ahzheng214@foxmail.com.}
\thanks{Jian Liang and Ran He are with National Laboratory of Pattern Recognition (NLPR) and the Center for Research on Intelligent Perception and Computing, CASIA. E-mail: liangjian92@gmail.com; rhe@nlpr.ia.ac.cn.}
}

\maketitle

\begin{abstract}
Unsupervised domain adaptation (UDA) aims to transfer knowledge from a labeled source domain to an unlabeled target domain.
Owing to privacy concerns and heavy data transmission, source-free UDA, exploiting the pre-trained source models instead of the raw source data for target learning, has been gaining popularity in recent years.
Some works attempt to recover unseen source domains with generative models, however introducing additional network parameters.
Other works propose to fine-tune the source model by pseudo labels, while noisy pseudo labels may misguide the decision boundary, leading to unsatisfied results.
To tackle these issues, we propose an effective method named Proxy-based Mixup training with label refinery (ProxyMix).
First of all, to avoid additional parameters and explore the information in the source model, ProxyMix defines the weights of the classifier as the class prototypes and then constructs a class-balanced proxy source domain by the nearest neighbors of the prototypes to bridge the unseen source domain and the target domain.
To improve the reliability of pseudo labels, we further propose the frequency-weighted aggregation strategy to generate soft pseudo labels for unlabeled target data. 
The proposed strategy exploits the internal structure of target features, pulls target features to their semantic neighbors, and increases the weights of low-frequency classes samples during gradient updating.
With the proxy domain and the reliable pseudo labels, we employ two kinds of mixup regularization, \textit{i.e.}, inter- and intra-domain mixup, in our framework, to align the proxy and the target domain, enforcing the consistency of predictions, thereby further mitigating the negative impacts of noisy labels.
Experiments on three 2D image and one 3D point cloud object recognition benchmarks demonstrate that ProxyMix yields state-of-the-art performance for source-free UDA tasks.
Code is available at \url{https://github.com/YuheD/ProxyMix}.
\end{abstract}


\IEEEpeerreviewmaketitle

\section{Introduction}

\begin{figure}
\center
\includegraphics[scale=0.25]{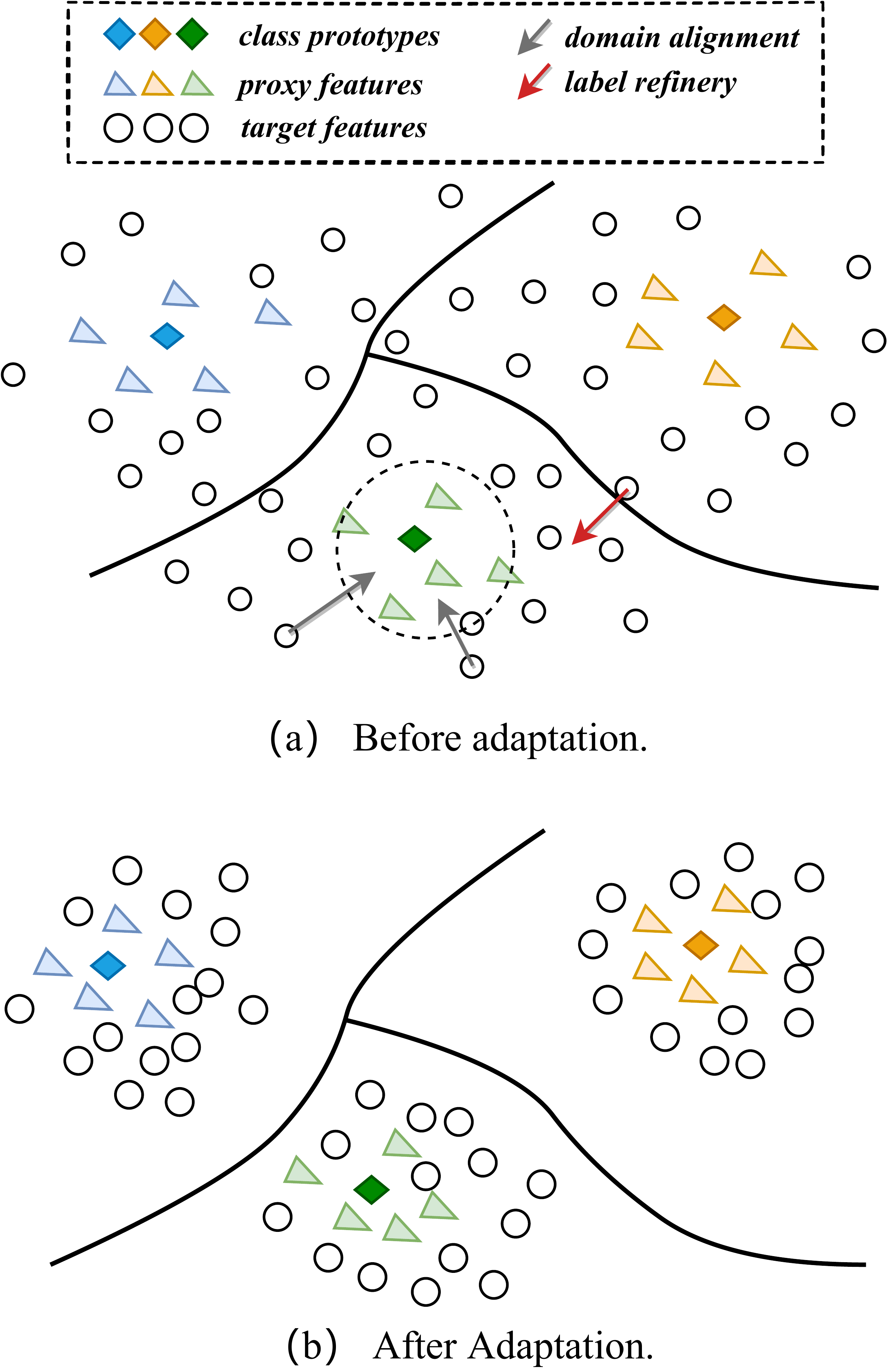}
\caption{The motivation of ProxyMix, which aligns the unseen source domain and target domain by two aspects: 1) aligning the proxy and target domain; and 2) refining the pseudo labels.}
\label{moti}
\end{figure}

The standard practice in the deep learning era---learning with massively labeled data---becomes expensive and laborious in many real-world scenarios.
Besides, the learned models often perform poorly in generalization to new unlabeled domains due to the domain discrepancy \cite{ben2007analysis}.
Hence, considerable efforts are devoted to unsupervised domain adaptation (UDA) \cite{dai2021disentangling,li2018domain,ganin2015unsupervised,long2018conditional}, which aims to transfer knowledge from a labeled source dataset to an unlabeled target dataset.
In recent years, UDA methods have been widely explored in various tasks such as image classification \cite{ganin2015unsupervised} and semantic segmentation \cite{tsai2018learning}.
The key problem of UDA is to alleviate the gap across different domains. Prior UDA methods mainly fall into three paradigms.
The first paradigm aims to pull the statistical moments of different feature distributions closer \cite{zellinger2017central,chen2020homm}, and the second paradigm introduces adversarial training with additional discriminators \cite{ganin2015unsupervised,tzeng2017adversarial}.
The last paradigm adopts various regularizations on the target network outputs like self-training or entropy-related objectives \cite{zou2018unsupervised,cui2020towards}.
Despite the impressive progress, the source data is always necessary during domain alignment, which might raise data privacy concerns nowadays.

The practical demand directly motivates a novel UDA setting named \emph{source-free domain adaptation} (SFDA) \cite{liang2020we,li2020model}, where only the well-trained source model instead of the well-annotated source dataset is provided to the target domain.
The booming efforts in the SFDA community are either generation-based or pseudo label-based.
The generation-based methods \cite{li2020model,tian2021vdm,qiu2021source} introduce extra generative modules to recover the unseen source domain at image-level or feature-level, and then address this problem from a UDA perspective.
Nevertheless, generative modules introduce additional parameters, and the recovered virtual source domain usually suffers from a mode collapse problem, which results in low-quality images or features.
The pseudo label-based methods \cite{qiu2021source,liang2021source,xia2021adaptive,huang2021model} label the target samples based on the present model's prediction or feature structure.
However, due to the extreme domain shift, the noises are inescapable, result in inaccurate decision boundary.

\begin{figure}
\center
\includegraphics[scale=0.5]{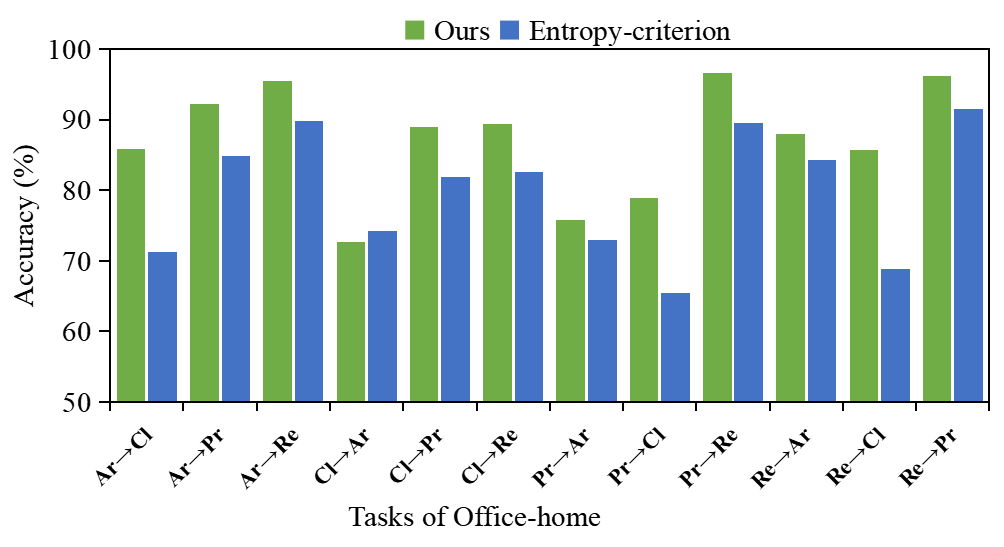}
\caption{The accuracies per task of proxy source domain on \textbf{Office-home}.}
\label{criterion}
\end{figure}

To address the issues above (additional parameters and noisy labels), we propose a new and effective method called Proxy-based Mixup training with label refinery (ProxyMix), to deal with the source-free domain adaptation problem.
To bridge the gap between the unseen source domain and the target domain while avoiding introducing extra parameters, we first select part of source-similar samples from the target domain rather than synthesize virtual images to construct a proxy source domain.
Specifically, we define the weights of the source classifier as the class prototypes \cite{papyan2020prevalence}, then select the nearest neighbors for each class prototype in angle space to construct the proxy source domain.
Priors methods with proxy source domain primarily employ entropy-criterion \cite{liang2021source,du2021generation}, which select samples with lower entropy for each class from pseudo-labeled target data.
In practice, as shown in Fig.~\ref{criterion}, we observe that the mean accuracy of our angle-induced proxy source domain is clearly higher than the entropy criterion.
Another significant benefit is that our pseudo labels are determined by the corresponding prototype, rather than the predictions from the source model, allowing us to create a class-balanced proxy source domain.

To improve the reliability of pseudo labels, we propose a frequency-weighted aggregation pseudo-labeling strategy (FA) as pseudo label refinery.
FA includes three operations applied to the predictions: sharpening, re-weighting, and aggregation.
Specifically, to avoid the ambiguous, we first sharpen the predictions of the classifier.
At the same time, we take the frequency of each class into account and re-weight the probability of each class, to improve the contribution of low-frequency classes and avoid bias to majority and easy classes in the target domain during gradient updating.
Then we introduce a non-parametric neighborhood aggregation strategy to pull the unlabeled target features close to their semantic neighbors, aiming to reduce the impact of outlier noisy labels and compact the semantic clusters.

With the proxy source domain, we tackle the challenging SFDA problem using a semi-supervised style with the aid of refined pseudo labels.
To align the proxy and target domain, while alleviating the negative consequence of noisy labels, two mixup regularizations \cite{zhang2017mixup, berthelot2019mixmatch, berthelot2020remixmatch, berthelot2021adamatch}, \textit{i.e.}, inter-domain and intra-domain mixup, are incorporated into our framework, enforcing the model to maintain consistency, thus improving the robustness against noisy labels.
As illustrated in Fig. \ref{moti}, the FA strategy refines the pseudo labels and compacts the feature clusters while the mixup training aligns the two domains, obtaining clear decision boundaries.

To summarize, the main contributions of this work are listed below in three-fold:
\begin{itemize}
\item We propose a simple yet effective method, ProxyMix, for source-free domain adaptation, which aims to discover a proxy source domain and utilize mixup training to implicitly bridge the gap between the target domain and the unseen source domain.
\item To obtain a reliable proxy source domain, we exploit the network weights of the source model and select source-like samples from the target domain in an efficient and accurate way.
\item To refine the noisy pseudo labels during alignment, we further propose a new frequency-weighted aggregation strategy, compacting the target feature clusters and avoiding bias to majority and easy classes. 
\end{itemize}

We conduct ablation study to verify the contribution and effectiveness of both proxy source domain construction and pseudo label refinery. Extensive results on four datasets further validate that ProxyMix yields comparable or superior performance to the state-of-the-art SFDA methods.

\section{Related Work}
\subsection{Unsupervised Domain Adaptation (UDA)}
UDA aims to transfer knowledge from a label-rich source domain to an unlabeled target domain.
UDA problems can be classified into four cases according to the relationship between the source and target domain, \textit{i.e.},  closed-set \cite{saenko2010adapting}, partial-set \cite{cao2018partial}, open-set \cite{panareda2017open}, and universal \cite{you2019universal}.
As a typical example of transfer learning, UDA provides methods to bridge domain gaps for various applications such as object recognition \cite{long2015learning, ganin2015unsupervised,dai2021disentangling,li2020maximum,li2018domain,xu2022few} and semantic segmentation \cite{tsai2018learning,zou2018unsupervised}.
The most prevailing paradigm for UDA is to extract domain-invariant features to align different domains while preserving the category information from the labeled source domain.
Roughly speaking, existing feature-level domain alignment could be divided into two different categories.
The first line \cite{ganin2015unsupervised,tzeng2017adversarial,long2018conditional} aligns representations by fooling a domain discriminator through adversarial training, while the second line \cite{long2015learning,sun2016deep} directly minimizes different discrepancy metrics (e.g., statistical moments) to match the feature distributions.
Besides, another line \cite{hoffman2018cycada} focuses on the image space alignment and converts the target image into a source style image (and visa versa).
By contrast, output-level regularization methods \cite{cui2020towards,jin2020minimum} achieve implicit domain alignment by forcing the target outputs to be diverse one-hot encodings.
\cite{liang2021domain} proposes an auxiliary classifier for target data to get the high-quality pseudo labels and \cite{liu2021cycle} introduces cycle self-training by utilizing target pseudo labels to train another head and enforce them to perform well on the source domain.
\cite{xu2020adversarial,wu2020dual} are the two most closely related works that introduce mixup training into adversarial UDA. However, our method does not require access to source data and develops a new pseudo label refinery strategy instead of focusing on the mix manner.

\subsection{Source-free Domain Adaptation (SFDA)} 
SFDA aims to tackle the domain adaptation problem without accessing the raw source data.
Before deep learning era, there are a number of transfer learning works \cite{yang2007cross,tommasi2010safety,kuzborskij2013stability,chidlovskii2016domain,liang2019distant} without source data that have been empirically successful.
In recent years, pioneering works \cite{liang2020we,li2020model} discover that the well-trained source model conceals sufficient source knowledge for the following target adaptation stage, and \cite{liang2020we} provides a clear definition of this problem.
The last two years have witnessed an increasing number of SFDA approaches \cite{qiu2021source,liang2021source,xia2021adaptive,huang2021model}, most of which are generation based \cite{li2020model,tian2021vdm,qiu2021source} or self-training \cite{liang2020we,yang2021exploiting} based methods.
Generation based methods \cite{tian2021vdm,qiu2021source,li2020model,yan2021source,du2021generation} generate virtual high-level features of the source domain to bridge the unseen source and target distribution.
Self-training based methods seek to refine the source model by using self-supervised techniques, with the pseudo label technique \cite{liang2020we,yang2021exploiting} being the most extensively employed.
\cite{xia2021adaptive,huang2021model} learn from target samples by distinct variants of contrastive learning.
\cite{yang2021exploiting} mines the hidden structure information such as the neighbor features to get the pseudo labels.
However, generating source samples usually introduces additional modules such as generators or discriminators, while pseudo-labeling might lead to wrong labels due to domain shift, both of which cause negative effects on the adaptation procedure.
Another practice \cite{yan2021source,du2021generation,liang2021source} is selecting part of the target data as a pseudo source domain, to compensate for the unseen source domain.
A typical method is entropy-criterion \cite{liang2021source}, which constructs the pseudo source domain by estimating a split ratio using the target dataset’s mean and maximum entropy, and then uses the split ratio to choose samples with lower entropy for all pseudo-labeled target domains within each class. The entropy-criterion provides a proxy source domain with a huge number of samples. However, the existence of hard classes and domain shift, causes the entropy-criterion to suffer from a severe class-imbalance problem.
Despite the fact that \cite{du2021generation} attempts to tackle this problem by simply choosing the same number for each class, there is no data in some hard classes, so the class-imbalance problem is unavoidable.
Unlike the previous works, our method builds the proxy source domain directly from the target domain using the source classifier weights, which is flexible and works well for SFDA.
Besides, our mixup training strategy is also different from theirs, which transfers the label information from the proxy source to the unlabeled target domain.

\subsection{Semi-Supervised Learning (SSL)}
SSL aims to combine supervised learning and unsupervised learning, leveraging the vast amount of unlabeled data with limited labeled data to improve the performance of classifier and to deal with the scenarios where labeled data is scarce \cite{van2020survey}.
As opposed to the domain adaptation problem, SSL deals with the samples from two identical domains.
SSL has flourished in recent years \cite{wang2020enaet, qin2021semi, li2019semi}, temporal ensemble \cite{laine2016temporal} introduces self-ensembling, forming a consensus prediction of the unknown labels using the outputs of the network-in-training on different epochs; 
MixMatch \cite{berthelot2019mixmatch} proposes a holistic approach for data-augmented unlabeled examples and mixing labeled and unlabeled data using mixup;
ReMixMatch \cite{berthelot2020remixmatch} aligns the distribution of labeled and unlabeled data.
FixMatch \cite{sohn2020fixmatch} demonstrates the strong performance of consistency regularizations and pseudo labels;
AdaMatch \cite{berthelot2021adamatch} proposes a unified approach to solve the unsupervised domain adaptation, semi-supervised learning, and semi-supervised domain adaptation problems.
Existing methods demonstrate the usefulness of mixup training in aligning distributions, and the growing popularity of SSL motivates us to convert the SFDA problem to an SSL challenge.
Such methods use the true labels, which are not available in our task, and these labels provide strong and diverse supervision.
Our data is pseudo-labeled, with little diversity and a lot of noise, so these semi-supervised learning approaches cannot be directly applied to our problem.

\begin{figure*}
\begin{center}
\includegraphics[scale=0.83]{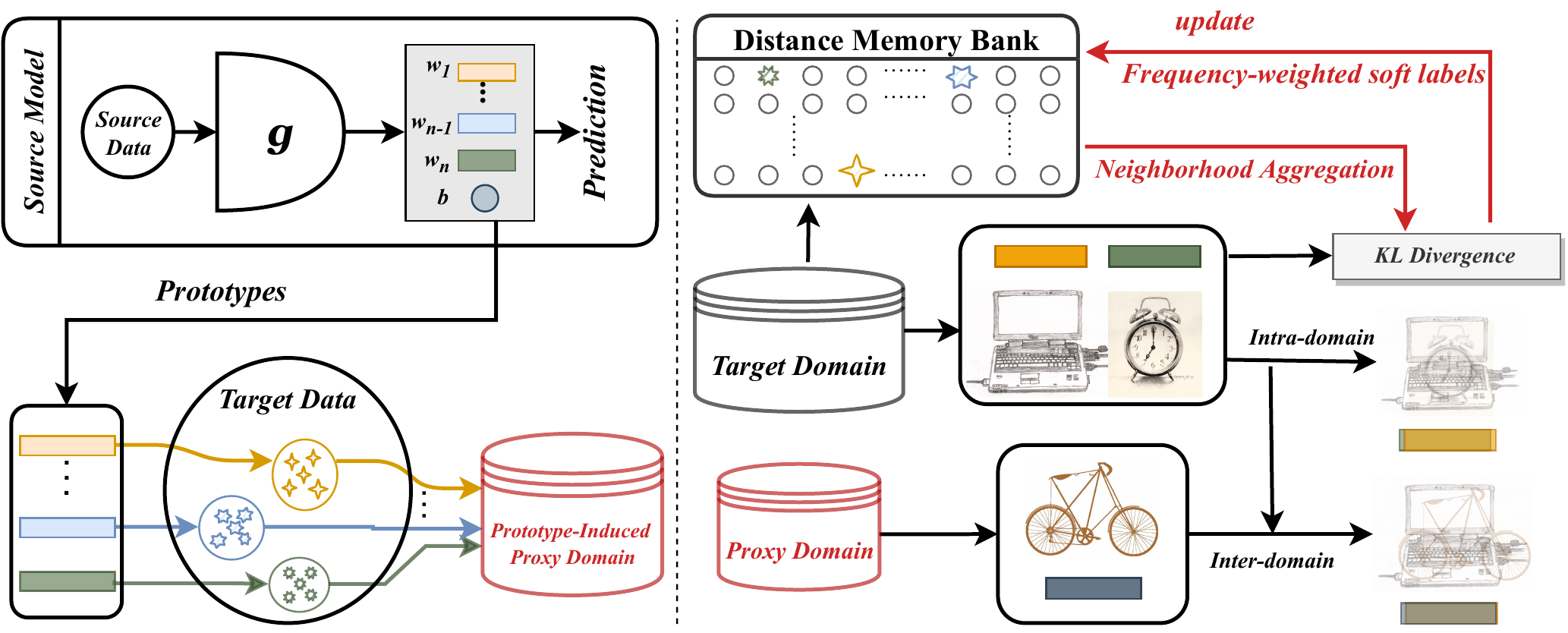}
\end{center}
\caption{Overview of ProxyMix on solving source-free domain adaptation. We treat the weights of the classifier as class prototypes to choose a series of confident samples to construct a class-balanced proxy source domain. Then the proxy source samples participate in two types of mixup training based on the proposed frequency-weighted soft label.}
\label{net-s1}
\end{figure*}

\section{Methodology} 
This paper mainly follows the problem definition of SHOT \cite{liang2020we} and focuses on a $K$-way visual classification task.
We aim to learn a target model $f_t:\mathcal{X}_t\to \mathcal{Y}_t$, and predict the label $y_t^i\in \mathcal{Y}_t$ for an input target image $x_t^i\in \mathcal{X}_t$ with only target data $\mathcal{X}_t$ and the well-trained source model $f_s:\mathcal{X}_s\to \mathcal{Y}_s$.
The model consists of two modules: the feature extractor $g:\mathcal{X}\to \mathbb{R}^d$ and the classifier $h:\mathbb{R}^d\to \mathbb{R}^K$.

Following the standard paradigm of SFDA \cite{liang2020we}, as a preliminary, we train the source model $f_s$ with the label smoothing \cite{LS} technique:
\begin{equation}
\begin{aligned}
&\mathcal{L}_{s r c}^{l s}\left(f_{s} ; \mathcal{X}_{s}, \mathcal{Y}_{s}\right)= \\
&\quad-\mathbb{E}_{\left(x_{s}, y_{s}\right) \in \mathcal{X}_{s} \times \mathcal{Y}_{s}} \sum_{k=1}^{K} l_{k}^{s} \log \delta_{k}\left(f_{s}\left(x_{s}\right)\right),
\end{aligned}
\end{equation}
where $l_{k}^{s}=(1-\alpha) q_k^{s}+\alpha/K$, $q^s$ is the one-hot encoding of $y_s$, $\alpha=0.1$ is the smoothing parameter, and $\delta_{k}(a)=\frac{\exp \left(a_{k}\right)}{\sum_{i} \exp \left(a_{i}\right)}$ is the soft-max output of the $K$-dimensional vector $a\in \mathbb{R}^K$.

During adaptation, we directly initialize the target model with the well-trained source model $f_t = f_s$, then freeze the classifier and fine-tune the feature extractor to ensure the target features are implicitly aligned with unseen source features via a same hypothesis.
It is worth noting that we do not adopt the special design of normalization techniques of SHOT \cite{liang2020we} for simplicity and commonality.

\subsection{Proxy Source Domain Construction by Prototypes}
\label{psd}
Recently, semi-supervised learning approaches \cite{berthelot2019mixmatch,berthelot2020remixmatch} have also shown impressive achievements on UDA problem, and Rukhovich et al. \cite{rukhovich2019mixmatchUDA} even wins the VisDA competition by directly exploiting MixMatch \cite{berthelot2019mixmatch} in 2019.
Inspired by them, we construct the proxy source domain by pseudo-labeling portions of confident samples (source-similar samples), and try to solve the SFDA task in a semi-supervised style.
Since the source data $\mathcal{X}_s$ is unavailable, we expect to mine the source information from the model $f_s$.
Previous works \cite{Tanwisuth2021prototype, yang2022we} leverage the weights of classifier as class prototypes in other fields, and obtain positive results.
Another classical practice \cite{papyan2020prevalence} exposes that the classifier weight vector of a well-trained last-layer classifier converges to a high-dimension geometry structure, which maximally separates the pair-wise angles of all classes in the classifier.
Therefore, inspired by these works, it is natural to select the nearest neighbors of classifiers' weights in angle space to construct the proxy source domain.
Concretely, we first define the weights $\{w_1,w_2,...w_K\}_{k=1}^{K}$ of the classifier $h_s$ as the class prototypes, where $K$ is the number of categories.
We use the class prototype $w_k$ as the cluster centroid to search and pseudo-label $N$ nearest samples in the unlabeled target domain $\mathcal{X}_t$ for the purpose of forming proxy source domain $\mathcal X_{ps}$: 
\begin{equation}
\begin{aligned}\label{xyps}
    & \{\mathcal{X}_{ps},\mathcal{Y}_{ps}\} = \{\mathcal{X}_{ps}^1,1\} \cup \cdots \cup \{\mathcal{X}_{ps}^K,K\}, \\
    \text{where}\; & \mathcal{X}_{ps}^k = \{x_{ps}; x_{ps}\in \mathop{{min}_N}\limits_{x_{t}} \left(\langle g_s(x_t),w_k \rangle \right) \},
\end{aligned} 
\end{equation}
and $\mathop{{min}_N}\limits_{x_{t}}(\cdot)_{k=1}^K$ denotes choosing $N$ samples $x_t$ with minimum distance for each class, $N$ is a hyper-parameter, deciding how many samples we select in each class.
To prevent the negative consequences caused by class imbalance, we select the same number of samples for each class.
$\langle a,b\rangle$ measures the distance between $a$ and $b$ in angle space, we use the cosine similarity by default. 
For these proxy source data, we directly calculate the cross entropy loss with labeling smoothing in the following,
\begin{equation} 
\begin{aligned} 
&\mathcal{L}_{ps}\left(f_{t} ; \mathcal{X}_{ps}, \mathcal{Y}_{ps}\right)= \\ 
&\quad-\mathbb{E}_{\left(x_{ps}, y_{ps}\right) \in \mathcal{X}_{ps} \times \mathcal{Y}_{ps}} \sum_{k=1}^{K} l_{k}^{ps} \log \delta_{k}\left(f_{t}\left(x_{ps}\right)\right),
\end{aligned} 
\end{equation} 
where $l_{k}^{ps}=(1-\alpha) q_k^{ps}+\alpha/K$ is the smoothed label, $q^{ps}$ denotes the one-hot encoding of $y_{ps}$.

\subsection{Pseudo-labeling by Frequency-weighted Aggregation (FA)}
Pseudo-labeling is a heuristic approach to semi-supervised learning, which progressively treats the predictions on unlabeled data as true labels, and often employs cross-entropy loss during training.
However, in an unsupervised learning setting, the class distribution is unknown, and the model is biased towards easy classes.
To mitigate the imbalance and sensitivity of pseudo labels, inspired by several classical works \cite{liang2021domain, xie2016unsupervised},
we propose a new pseudo label refinery strategy to get reliable soft pseudo labels in the presence of domain shift.
In specific, we adjust the class distribution of the prediction to alleviate the class imbalance, and then we use the center of semantic neighbors as the pseudo label, rather than depending on a single prediction.
This compacts the cluster by pulling the unlabeled target features closer to their semantic neighbors, resulting in a clear classification boundary.
Note that hard labels reinforce the confidence of the current model, while losing some information. Hence we use the soft predictions rather than the one-hot vectors as the pseudo labels, which are able to provide more distribution information and decrease the negative effect of corrupted one-hot labels.

\begin{figure}
\center
\includegraphics[scale=0.235,trim={0.0cm 0.0cm 4.0cm 0.0cm}]{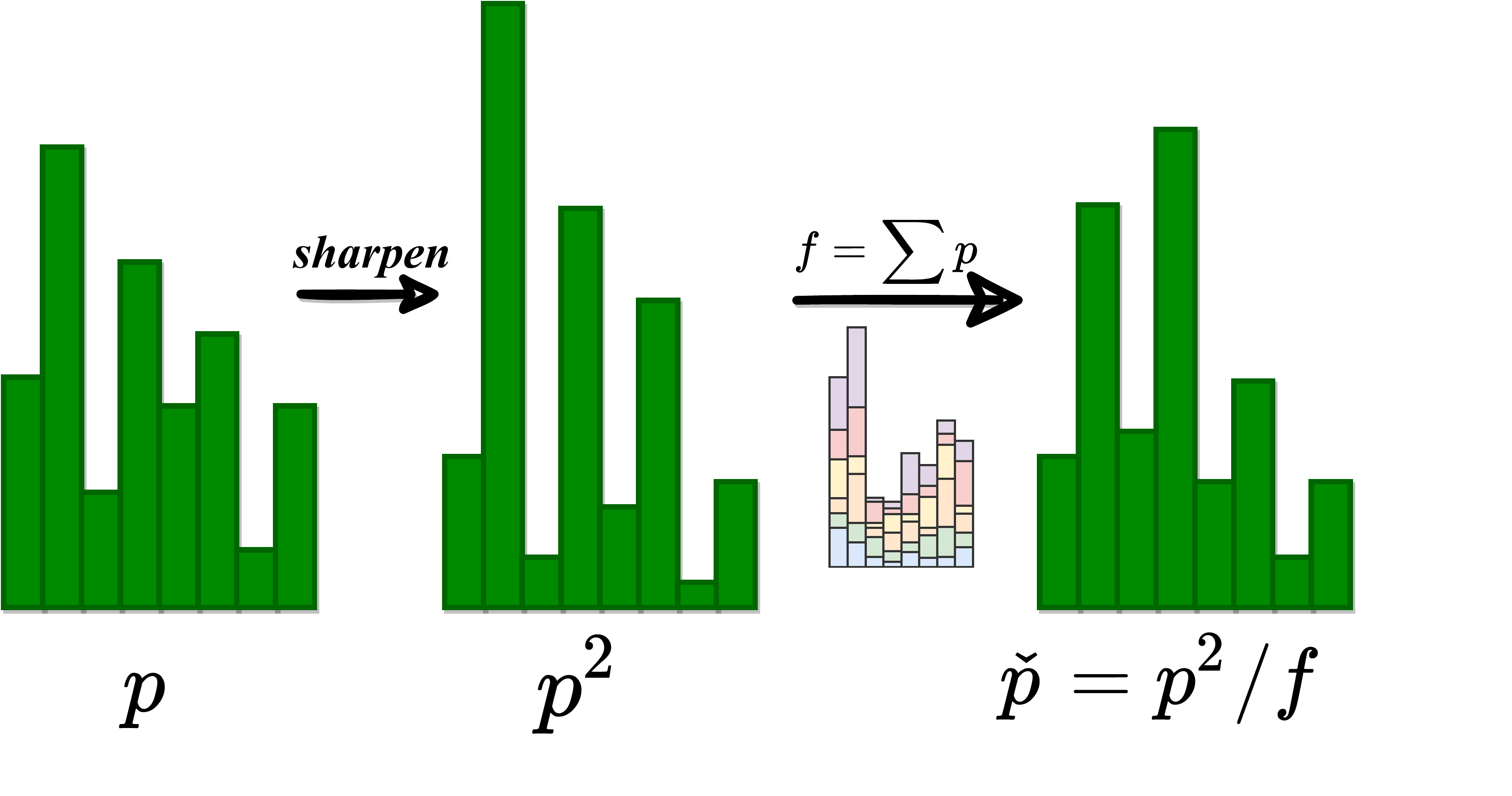}
\caption{ Illustration of the frequency-weighted strategy as label refinery. We first sharpen the predictions to the second power, and then normalize the predictions by the frequency per class. 
}
\label{fw}
\end{figure}

\noindent
\textbf{Neighborhood Aggregation.}
To leverage the local data structure, we employ the neighborhood aggregation strategy, which is based on the idea of message passing via neighbors, to adjust the predictions of the input target data.
Concretely, we construct a large memory bank to store both the features and the predictions of target data.
During pseudo-labeling, we retrieve $m$ nearest neighbors from the memory bank for each sample in the current mini-batch according to their features $g_t(x_t^i)$, and calculate the soft label $\hat{q}_{i}$ of data point $x_t^i$ by aggregating these predictions of feature-level neighbors: 

\begin{equation}\label{softlabel}
\hat{q}_{i}=\frac{1}{m} \sum_{j \neq i, j \in \mathcal{N}_{i}} \check{p}_{j},
\end{equation}
where $\mathcal{N}_i$ is the neighbor index set of the data $x_t^i$, $\check{p}_{j}$ are the frequency-weighted predictions of neighbors stored in the bank, then we explain how these predictions are obtained.

\noindent
\textbf{Frequency-weighted prediction.}
As illustrated in Fig.~\ref{fw}, to avoid ambiguity, we first sharpen the calculated output predictions $p_i$.
Besides, the network will be empirically skewed towards these majority classes due to the class imbalance.
Then, we further multiply the predictions by a weight based on the frequency of the class.
In specific, given the soft-max output predictions $p_i=\delta (f_t(x_t^i))$, the frequency-weighted predictions can be obtained through  
\begin{equation}
\{\check{p}_{ij}\}_{j=1}^K=\left\{\frac{p_{i j}^{2} / f_{j}}{\sum_{j^{\prime}}(p_{i j^{\prime}}^{2} / f_{j^{\prime}})}\right\}_{j=1}^K,
\end{equation}
where $f_{j}=\sum_{i} p_{i j}$ are soft cluster frequencies calculated by the current batch of samples.
Through the operation above, we expect to achieve class-balance in the predictions. 
At each iteration, we update the features and predictions associated with the data in the corresponding location in the memory bank.

\subsection{Domain Alignment by Mixup Training}
Two mixup training procedures are incorporated in our method.
In essence, mixup trains a neural network on convex combinations of pairs of examples and their labels to regularize the network to support linear behavior in-between training samples.
Pioneers have proved the effectiveness of mixup training on UDA and SSL tasks \cite{zhang2017mixup, berthelot2019mixmatch,berthelot2020remixmatch,rukhovich2019mixmatchUDA}.
Such a simple regularization can improve the generalization and the robustness to some noisy labels, so it is suitable for pseudo label-based unsupervised learning tasks.
Inspired by these methods, with the prototype-induced pseudo source domain $\{\mathcal{X}_{ps}, \mathcal{Y}_{ps}\}$ and target domain $\mathcal{X}_t$, we introduce two different regularizations via mixup training.

\noindent
\textbf{Inter-domain Mixup.} 
To align the proxy source domain and the target domain, we employ inter-domain mixup regularization.
\cite{berthelot2019mixmatch} mixes the labeled data with both unlabeled data and labeled data itself.
However, the ``labeled'' data in our case is not completely trustworthy. 
As a result, we do not add any mixup training between the proxy source samples, but only between the pseudo source domain and the target domain only, constructing in virtual training samples below:
\begin{equation} 
\begin{aligned} 
&\tilde{x}_{r}=\rho x_{ps}+(1-\rho) x_{t}, \\
&\tilde{q}_{r}=\rho q_{ps}+(1-\rho) \hat q, 
\notag
\end{aligned} 
\end{equation}
where $q_{ps}$ denotes the one-hot encoding of $y_{ps}$, and $\hat q$ is the soft label of $x_t$ calculated by Eq. (\ref{softlabel}), $\rho$ is the mixup coefficient sampled from a random Beta distribution.

Then we adopt the KL divergence to calculate the soft label classification loss:
\begin{equation}\label{inter}
\begin{aligned}
\mathcal{L}_{tgt}^{inter} 
&=\mathrm{KL}( \tilde{q}_{r} \| \delta (f_t(\tilde{x}_{r}))).
\end{aligned}
\end{equation}

\begin{algorithm}
	\renewcommand{\algorithmicrequire}{\textbf{Input:}}
	\renewcommand{\algorithmicensure}{\textbf{Output:}}
	\caption{Algorithm of the proposed ProxyMix.}
	\label{alg1}
	\begin{algorithmic}[1]
	    \REQUIRE Target dataset $\mathcal{X}_t$; well-trained source model $f=g\cdot h$, where $g:\mathcal{X}\to \mathbb{R}^d$ is the feature extractor and $h:\mathbb{R}^d\to \mathbb{R}^K$ is the classifier; 
		\STATE Build the proxy source domain $\{ \mathcal{X}_{ps}, \mathcal{Y}_{ps}\}$ by Eq. (\ref{xyps});
		\STATE Initialize the feature memory bank $B_f$ and prediction memory bank $B_l$;
		\REPEAT
		\STATE Randomly sample a batch of target data $x_t$ from $\mathcal{X}_t$ and proxy source data $x_{ps}$ from $\mathcal{X}_{ps}$;
		\STATE Obtain the soft label $\hat q$ of $x_t$ by Eq. (\ref{softlabel});
		\STATE Update $g$ by Eq. (\ref{total});
		\STATE Update the corresponding features and predictions of $x_t$ in feature bank $B_f$ and prediction bank $B_l$;
		\UNTIL Iterations are exhausted.
		\ENSURE New model $f=g\cdot h$.
	\end{algorithmic}  
\label{algo}
\end{algorithm}

\setlength{\tabcolsep}{3.0pt}
\begin{table*}
	\centering
	\caption{Classification accuracies (\%) of state-of-the-art methods on  \textbf{Office-home} \protect\cite{home} (ResNet-50). SF denotes source-free. We use \textbf{Bold} to highlight the best and \underline{underline} to highlight the second best among source-free methods.}
	\resizebox{0.95\textwidth}{!}{$	
	\begin{tabular}{llccccccccccccc}
		\toprule
		SF& Method  &Ar$\to$Cl & Ar$\to$Pr & Ar$\to$Re & Cl$\to$Ar & Cl$\to$Pr & Cl$\to$Re & Pr$\to$Ar & Pr$\to$Cl & Pr$\to$Re & Re$\to$Ar & Re$\to$Cl & Re$\to$Pr & Avg. \\
		\midrule
		&No Adapt.   & 46.1 &	67.0 &	74.3 &	52.0 &	62.7 &	64.3 &	53.8 &	42.1 &	73.7 &	67.0 &	47.7 &	78.2 &	60.7  \\
		\midrule
		$\times$& MCD \cite{saito2018maximum} & 48.9 & 68.3 & 74.6 & 61.3 & 67.6 & 68.8 & 57.0 & 47.1 & 75.1 & 69.1 & 52.2 & 79.6 & 64.1 \\
		$\times$& CDAN \cite{long2018conditional} & 50.7 & 70.6 & 76.0 & 57.6 & 70.0 & 70.0 & 57.4 & 50.9 & 77.3 & 70.9 & 56.7 & 81.6 & 65.8\\
		$\times$& SAFN \cite{xu2019larger} &52.0& 71.7& 76.3 &64.2& 69.9 &71.9 &63.7 &51.4& 77.1 &70.9& 57.1 &81.5 &67.3\\
		$\times$& SymNets \cite{zhang2019domain} & 47.7& 72.9 &78.5& 64.2 &71.3& 74.2& 64.2& 48.8 &79.5& 74.5 &52.6& 82.7& 67.6 \\
		$\times$& MDD \cite{zhang2019bridging} & 54.9& 73.7 &77.8 &60.0 &71.4& 71.8& 61.2& 53.6& 78.1& 72.5& 60.2& 82.3& 68.1 \\
		$\times$& TADA \cite{wang2019transferable} & 53.1 &72.3 &77.2 &59.1& 71.2& 72.1& 59.7& 53.1& 78.4& 72.4& 60.0 &82.9& 67.6 \\
		$\times$& BNM \cite{cui2020towards}&52.3 &73.9& 80.0& 63.3& 72.9 &74.9& 61.7& 49.5& 79.7& 70.5& 53.6& 82.2 &67.9 \\
		$\times$& BDG \cite{yang2020bi} & 51.5 &73.4& 78.7& 65.3& 71.5& 73.7& 65.1 &49.7 &81.1& 74.6& 55.1& 84.8& 68.7 \\
		$\times$& SRDC \cite{tang2020unsupervised} & 52.3& 76.3& 81.0 &69.5& 76.2& 78.0& 68.7& 53.8& 81.7& 76.3& 57.1 &85.0 &71.3 \\
		$\times$& RSDA-MSTN \cite{gu2020spherical} & 53.2 &77.7& 81.3& 66.4& 74.0 &76.5& 67.9& 53.0& 82.0& 75.8& 57.8 &85.4& 70.9 \\
		$\times$& ATDOC \cite{liang2021domain} &60.2& 77.8& 82.2& 68.5 &78.6& 77.9& 68.4& 58.4& 83.1& 74.8& 61.5 &87.2 &73.2 \\
		\midrule
		\checkmark& SSFT-SSD \cite{yan2021source} & 51.7&76.0&	79.9&	66.8&	75.8&	77.2&	63.9&	52.1&	80.6&	73.5&	57.1&	83.0	&69.8 \\
		\checkmark&  VDM-DA \cite{tian2021vdm}  &\underline{\textbf{59.3}} &75.3 &	78.3 &	67.6 &	76.0 &	75.9 &	\textbf{68.8} &	57.7 &	79.6 &	74.0 &	61.1 &	83.6 &	71.4   \\
		\checkmark& CPGA \cite{qiu2021source}  & \underline{\textbf{59.3}} &78.1& 79.8& 65.4& 75.5 &76.4& 65.7& \textbf{58.0} &81.0& 72.0& \textbf{64.4}& 83.3& 71.6 \\
		\checkmark& SHOT \cite{liang2020we}  & 57.1 &	78.1 &	81.5 &	68.0 &	78.2 &	78.1 &	67.4 &	54.9 &	82.2 &	73.3 &	58.8 &	84.3 &	71.8   \\
		\checkmark& PS \cite{du2021generation} & 57.8&	77.3&	81.2&	\textbf{68.4}&	76.9&	78.1&	67.8&	57.3&	82.1&	\textbf{75.2}&	59.1&	83.4&	72.1 \\
		\checkmark& NRC \cite{yang2021exploiting}  & 57.7& \underline{80.3} &82.0 & \underline{68.1} &\textbf{79.8}& \underline{78.6} & 65.3 &56.4& \textbf{83.0}& 71.0& 58.6 &\underline{\textbf{85.6}} &72.2 \\
		\checkmark&  $\textrm{A}^2\textrm{Net}$ \cite{xia2021adaptive} &58.4 &	79.0 &	\textbf{82.4} &	67.5 &	79.3 &	\textbf{78.9} &	\underline{68.0} &	56.2 &	\underline{82.9} &	\underline{74.1} &	60.5 &	85.0 &	\underline{\textbf{72.8}}   \\
		\checkmark&  ProxyMix  &\underline{\textbf{59.3}} &	\textbf{81.0} &	\underline{81.6} &	65.8 &	\underline{79.7} &	78.1 &	67.0 &	\underline{57.5} &	82.7 &	73.1 &	\underline{61.7} &	\underline{\textbf{85.6}} &	\underline{\textbf{72.8}}  \\
		\bottomrule
	\end{tabular}
	$}
	\label{tab:home}
\end{table*}

\setlength{\tabcolsep}{7.0pt}
\begin{table*}
	\centering
	\caption{Classification accuracies (\%) on \textbf{Office-31} \protect\cite{office31} (ResNet-50). [$*$: mean values except D$\leftrightarrow$W.]}
    \resizebox{0.75\textwidth}{!}{$
    	\begin{tabular}{llcccccccc}
    		\toprule
    		SF&Method &  A$\to$D & A$\to$W & D$\to$A & D$\to$W & W$\to$A & W$\to$D & Avg. & $\textrm{Avg}.^*$ \\
    		\midrule
    		&No Adapt. & 77.3 &	73.8 &	59.9 &	96.5 &	60.7 &	98.4 &	77.8  & 67.9 \\
    		\midrule
    		$\times$ &MCD \cite{saito2018maximum} &92.2 &88.6 &	69.5 &	98.5 &	69.7 &	100.0  &86.5 & 80.0 \\
    		$\times$ &CDAN \cite{long2018conditional} &92.9 &	94.1 &	71.0 &	98.6 &	69.3 &	100.0  &87.7 & 81.8\\
    		$\times$ &MDD \cite{zhang2019bridging} &90.4 &	90.4 &	75.0 &	98.7 &	73.7 &	99.9 & 88.0 & 82.4 \\
    		$\times$ &BNM \cite{cui2020towards} &90.3 &	91.5 &	70.9 &	98.5 &	71.6 &	100.0  &87.1 & 81.1 \\
    		$\times$&DMRL \cite{wu2020dual} & 93.4 &	90.8 &	73.0 &	99.0 &	71.2 &	100.0  &87.9 & 82.1 \\
    		$\times$&BDG \cite{yang2020bi} & 93.6 &	93.6 &	73.2 &	99.0 &	72.0 &	100.0  &88.5 & 83.1 \\
    		$\times$ &MCC \cite{jin2020minimum} &95.6 &	95.4 &	72.6 &	98.6 &	73.9 &	100.0  &89.4 & 84.4 \\
    		$\times$ &SRDC \cite{tang2020unsupervised} &95.8 &	95.7 &	76.7 &	99.2 &	77.1 &	100.0 & 90.8 & 86.3 \\
    		$\times$ &RWOT \cite{xu2020reliable} &94.5 	&95.1 &	77.5 &	99.5 &	77.9 &	100.0  &90.8 & 86.3 \\
    		$\times$ &RSDA-MSTN \cite{gu2020spherical}& 95.8 &	96.1 &	77.4 &	99.3 &	78.9 &	100.0  &91.1 & 87.1 \\
    		$\times$ &ATDOC \cite{liang2021domain}&  95.4& 94.6& 77.5 &98.1& 77.0& 99.7 & 90.4& 86.1 \\
    		\midrule
    		\checkmark &SHOT \cite{liang2020we}& 94.0 &	90.1 &	74.7 &	98.4 &	74.3 &	99.9 &	88.6 & 83.3  \\
    		\checkmark  &SSFT-SSD \cite{yan2021source}&95.2& \underline{95.0} &72.7 &\underline{98.7} &73.5& \underline{\textbf{100.0}} &89.2 & 84.1  \\
    		\checkmark  &NRC \cite{yang2021exploiting}& \textbf{96.0} &	90.8 &	75.3 &	\textbf{99.0} &	75.0 &	\underline{\textbf{100.0}} &	89.4 & 84.3  \\
    		\checkmark  &HCL \cite{huang2021model}& 94.7 &	92.5 &	\underline{75.9} &	98.2 &	\textbf{77.7} & \underline{\textbf{100.0}} &	89.8  & 85.2 \\
    		\checkmark  &CPGA \cite{qiu2021source}& 94.4 &	94.1 &	\textbf{76.0} &	98.4 &	\underline{76.6} &	99.8 &	\underline{89.9} & \underline{85.3}  \\
    		\checkmark  &ProxyMix& \underline{95.4} &	\textbf{96.7} &	75.1 &	98.5 &	75.4 &	99.8 &	\textbf{90.1} & \textbf{85.6} \\
    		\bottomrule
    	\end{tabular}
	$}
	\label{tab:office}
\end{table*}

\noindent
\textbf{Intra-domain Mixup.} 
To mine the inner structure of the target domain, we also adopt the mixup regularization between different target data.
As it is typical in many SSL methods, we use data augmentation on target data.
In specific, for each mini-batch of target data $x_t$, we concatenate it with its augmented version $\hat x_t$ to construct a vector notated as $x_a = \mathrm{cat}(x_t,\hat x_t)$.
Then we mixup $x_a$ and its shuffled version $x_{a}^s$ to construct the virtual training samples below: 
\begin{equation}
\begin{aligned} 
&\tilde{x}_{a}=\rho x_{a}+(1-\rho) x_{a}^s, \\ 
&\tilde{q}_{a}=\rho \hat q_{a}+(1-\rho) \hat q_a^s,
\end{aligned} 
\notag
\end{equation}
where $x_{a}^s$ is the shuffled version of $x_a$, $\hat q_a$ and $\hat q_a^s$ are the soft label of $x_a$ and $x_a^s$ calculated by Eq. (\ref{softlabel}), respectively.
Then we formulate the intra-domain mixup regression loss as:
\begin{equation} 
\mathcal{L}_{tgt}^{intra} = \|f_t(\tilde{x}_{a}) -\tilde{q}_{a}\|_2^2.
\end{equation}
Note here we use square $L_2$ loss. Unlike the cross entropy loss used in Eq. (\ref{inter}), it is bounded and more robust due to the insensitivity to corrupted labels.

\setlength{\tabcolsep}{5.0pt}
\begin{table*}
\centering
\caption{Classification accuracies (\%) on the large-scale synthesized-to-real dataset \textbf{VisDA} \protect\cite{peng2017visda} (ResNet-101).}
\resizebox{0.95\textwidth}{!}{$
\begin{tabular}{llccccccccccccc}
\toprule
SF &Method & plane& bicycle& bus& car& horse& knife &mcycl& person& plant& sktbrd &train& truck& Per-class \\
\midrule
& No Adapt. & 63.2 &	10.4 &	47.6 &	73.0 &	46.9 &	4.5 &	66.4 &	15.6 &	62.1 &	17.7 &	88.5 &	7.2 &	41.9    \\
\midrule
$\times$& ADR \cite{saito2017adversarial} &94.2& 48.5 &84.0& 72.9 &90.1 &74.2& 92.6 &72.5 &80.8 &61.8 &82.2 &28.8& 73.5 \\
$\times$&CDAN \cite{long2018conditional} & 85.2 &66.9 &83.0& 50.8& 84.2 &74.9 &88.1& 74.5 &83.4 &76.0 &81.9 &38.0 &73.9 \\
$\times$&CDAN+BSP \cite{chen2019transferability} & 92.4& 61.0 &81.0 &57.5 &89.0 &80.6 &90.1 &77.0 &84.2 &77.9 &82.1 &38.4 &75.9 \\
$\times$&SAFN \cite{xu2019larger} & 93.6& 61.3 &84.1 &70.6 &94.1 &79.0 &91.8 &79.6 &89.9 &55.6 &89.0 &24.4 &76.1 \\
$\times$&SWD \cite{lee2019sliced} & 90.8 &82.5& 81.7& 70.5 &91.7& 69.5& 86.3 &77.5 &87.4 &63.6& 85.6 &29.2 &76.4 \\
$\times$&MDD \cite{zhang2019bridging} & - &-& -& -& -& -& -& -& -& -& -& -& 74.6 \\
$\times$&DMRL \cite{wu2020dual} & - &- &- &- &- &- &- &- &- &- &- &- &75.5 \\
$\times$&MCC \cite{jin2020minimum} & 88.7& 80.3 &80.5 &71.5& 90.1 &93.2& 85.0& 71.6 &89.4& 73.8 &85.0& 36.9& 78.8 \\
$\times$& STAR \cite{lu2020stochastic} &95.0 &84.0 &84.6 &73.0 &91.6 &91.8& 85.9 &78.4& 94.4& 84.7& 87.0 &42.2 &82.7\\
$\times$&RWOT \cite{xu2020reliable} & 95.1 &80.3 &83.7 &90.0 &92.4& 68.0 &92.5 &82.2 &87.9& 78.4& 90.4 &68.2 &84.0\\
$\times $&ATDOC \cite{liang2021domain} &  93.0& 77.4& 83.4 &62.3& 91.5 &88.4 &91.8& 77.1& 90.9& 86.4 &85.8 &48.2& 81.4 \\
\midrule
\checkmark &SSFT-SSD \cite{yan2021source} &  95.4 &86.5 &79.3& 51.5& 92.9& 94.5& 82.1& 79.7& 90.0 &87.1& 87.8& 57.9 &82.1 \\
\checkmark  &SHOT \cite{liang2020we} &  94.3& 88.5 &80.1& 57.3& 93.1& 94.9& 80.7& 80.3 &91.5& 89.1 &86.3 &58.2 &82.9 \\
\checkmark  &HCL \cite{huang2021model} &  93.3& 85.4& 80.7& 68.5& 91.0 &88.1& 86.0 &78.6 &86.6 &88.8 &80.0& \textbf{74.7}& 83.5 \\
\checkmark  &PS \cite{du2021generation} &  95.3&	86.2&	82.3&	61.6&	93.3&	95.7&	86.7&	80.4&	91.6&	90.9&	86.0&	59.5&	84.1 \\
\checkmark  &$\textrm{A}^2\textrm{Net}$ \cite{xia2021adaptive} &  94.0 &87.8& \underline{85.6}& 66.8& 93.7& 95.1& 85.8& 81.2& 91.6& 88.2& 86.5 &56.0 &84.3 \\
\checkmark &VDM-DA \cite{tian2021vdm} &  \textbf{96.9}&	\underline{89.1}&	79.1&	66.5&	\underline{95.7}&	\underline{96.8}&	85.4&	83.3&	96&	86.6&	89.5&	56.3&	85.1\\
\checkmark  &NRC \cite{yang2021exploiting} &  \underline{96.8} & \textbf{91.3} &82.4& 62.4& \textbf{96.2} &95.9 &86.1 &80.6 &\textbf{94.8}& \textbf{94.1}& \textbf{90.4}& 59.7& \underline{85.9} \\
\checkmark  &CPGA \cite{qiu2021source} &  95.6& 89.0 &75.4 &64.9& 91.7&\textbf{97.5}& \underline{89.7}& \underline{83.8} & \underline{93.9}& \underline{93.4}& 87.7 &\underline{69.0}& \textbf{86.0} \\
\checkmark  & ProxyMix & 95.4 &	81.7 &	\textbf{87.2} &	\textbf{79.9} &	95.6 &	\underline{96.8} &	\textbf{92.1} &	\textbf{85.1} &	93.4 &	90.3 &	\underline{89.1} &	42.2 &	85.7   \\
\bottomrule
\end{tabular} 
$}
\label{tab:visda} 
\end{table*} 

\setlength{\tabcolsep}{5.0pt}
\begin{table*}
\centering
\caption{Classification accuracies (\%) on the 3D point cloud dataset \textbf{PointDA-10 \cite{qi2017pointnet}} (PointNet \cite{qin2019pointdan}). The results except ours are from NRC \cite{yang2021exploiting} and PointDAN \cite{qin2019pointdan}.}
\resizebox{0.7\textwidth}{!}{$
 \begin{tabular}{llccccccc}
 \toprule
 SF &Method  &  M $\rightarrow$ S & M$\rightarrow$ $\textrm{S}^*$ & S $\rightarrow$ M & S $\rightarrow$ $\textrm{S}^*$ & $\textrm{S}^*$ $\rightarrow$ M & $\textrm{S}^*$ $\rightarrow$ S & Avg. \\
 \midrule
 &No Adapt. &21.5&	21.7&	18.5&	29.5&	18.8&	25.8&	22.6\\
 \midrule
 $\times$ &MMD \cite{long2013transfer} &  57.5 & 27.9 & 40.7 & 26.7 & 47.3 & 54.8 & 42.5 \\
 $\times$ &DANN \cite{ganin2015unsupervised} &  58.7 & 29.4 & 42.3 & 30.5 & 48.1 & 56.7 & 44.2\\
 $\times$ &ADDA \cite{tzeng2017adversarial} &  61.0 & 30.5 & 40.4 & 29.3 & 48.9 & 51.1 & 43.5\\
 $\times$ &MCD \cite{saito2018maximum} &  62.0 & 31.0 & 41.4 & 31.3 & 46.8 & 59.3 & 45.3 \\
 $\times$ &PointDAN \cite{qin2019pointdan} &  64.2 & 33.0 & 47.6 & 33.9 & 49.1 & 64.1 & 48.7\\
 \midrule
 \checkmark &VDM-DA \cite{tian2021vdm} &  58.4& \textbf{30.9} &\textbf{61.0}& \textbf{40.8}& 45.3& 61.8& 49.7 \\
 \checkmark &NRC \cite{yang2021exploiting} &  \underline{64.8} & \underline{25.8} & 59.8 & 26.9 & \underline{70.1} & \textbf{68.1} & \underline{52.6} \\
 \checkmark &ProxyMix & \textbf{65.2} & 22.4 & \underline{60.8} & \underline{30.8} & \textbf{81.2} & \underline{64.2} & \textbf{54.1} \\
 \bottomrule
 \end{tabular} 
$}
\label{tab:pointda}
\end{table*}

\subsection{Overall Objective}
Combining the proxy source classification loss and two types of mixup loss, our overall objective is formulated as:

\begin{equation}\label{total}
\mathcal{L}_{total} = \mathcal{L}_{ps} + \lambda\mathcal{L}_{tgt}^{inter} +\eta\mathcal{L}_{tgt}^{intra} 
\end{equation} 
where $\lambda$ and $\eta$ are trade-off parameters to balance losses. 
The overall pipeline of ProxyMix is illustrated in Algorithm \ref{algo}.

\section{Experiment}
\noindent
\textbf{Datasets.}
We conduct the experiments on four popular benchmark datasets:
(1) \textbf{Office-31} \cite{office31} is a standard domain adaptation dataset consisting of three distinct domains, \textit{i.e.,} Amazon (A), DSLR (D) and Webcam (W), and 31 categories in the shared label space.
The specific numbers of images for each domain are 2,817 (A), 498 (D) and 795 (W), therefore the dataset suffers from severe data imbalance.
(2) \textbf{Office-home} \cite{home} is a medium-sized domain adaptation dataset with 15,500 images collected from four domains Art (Ar), Clipart (Cl), Product (Pr), and Real-World (Re).
There are 65 categories per domain, which is much more than \textbf{Office-31}.
(3) \textbf{VisDA} \cite{peng2017visda} is a large-scale challenging dataset which consists of a 12-class synthesize-to-real object recognition task.
The source domain involves 152k synthetic images which are produced by 3D rendering model under various conditions.
The target domain contains 55k images collected from the real-world scene.
(4) \textbf{PointDA-10} \cite{qin2019pointdan} is a common-used 3D cloud-point dataset extracted from three popular 3D object/scene datasets, \textit{i.e,} modelnet (M) shapenet (S), and scannet ($\textrm{S}^*$) for cross-domain 3D object recognization. Each domain contains its own training and testing sets. We train our models by source and target domain's training set, and show the test resutls on the target domain's test set.

\noindent
\textbf{Baselines.}
We compare ProxyMix with the state-of-the-art source-free domain adaptation methods: 
SHOT \cite{liang2020we}, CPGA \cite{qiu2021source}, $\textrm{A}^2\textrm{Net}$ \cite{xia2021adaptive}, HCL \cite{huang2021model}, NRC \cite{yang2021exploiting}, SSFT-SSD \cite{yan2021source}, PS \cite{du2021generation}.
Moreover, to illustrate the effectiveness of ProxyMix, we further compare our method with the state-of-the-art UDA methods:
SymNets \cite{zhang2019domain}, TADA \cite{wang2019transferable}, BNM \cite{cui2020towards}, BDG \cite{yang2020bi}, SRDC \cite{tang2020unsupervised}, RSDA-MSTN \cite{gu2020spherical}, ADR \cite{saito2017adversarial}, CDAN \cite{long2018conditional}, CDAN+BSP \cite{chen2019transferability}, SAFN \cite{xu2019larger}, SWD \cite{lee2019sliced}, MDD \cite{zhang2019bridging}, DMRL \cite{wu2020dual}, MCC \cite{jin2020minimum}, STAR \cite{lu2020stochastic}, RWOT \cite{xu2020reliable}, ATDOC \cite{liang2021domain}, MMD \cite{long2013transfer}, DANN \cite{ganin2015unsupervised},  ADDA \cite{tzeng2017adversarial},  MCD \cite{saito2018maximum}, PointDAN \cite{qin2019pointdan}.
We use \textbf{bold} to highlight the best results and \underline{underline} to highlight the second best results among \emph{source-free methods}.

\noindent
\textbf{Implementation Details.}
We implement our method based on PyTorch.
For network architecture, we adopt ResNet \cite{resnet}, pretrained on the ImageNet as the backbone, and replace the original fully connected layer with a bottleneck layer followed by a task-specific linear layer.
In the source model training stage, we exploit SGD optimizer with learning rate $1e^{-3}$ for backbone and $1e^{-2}$ for the bottleneck and classifier.
In the target adaptation stage, we use SGD optimizer with learning rate $1e^{-3}$ for the backbone and freeze the fully connected classification layer.
The numbers of epoch are set to 30, 50, 5 in training stage and 50, 50, 1 in adaptation stage for \textbf{Office-31}, \textbf{Office-home} and \textbf{VisDA}, respectively.
Specially, for \textbf{PointDA-10}, we follow the open source code of NRC \cite{yang2021exploiting}, use PointNet \cite{qi2017pointnet} as our backbone network, learning rate $1e^{-6}$ and Adam optimizer with 100 epochs each stage.
For the hyper-parameters, considering the confidence of pseudo labels, we set $\lambda = 1 $, $\eta =100 $, and we alter $\lambda$ and $\eta$ linearly by multiplying a ratio that varies linearly from 0 to 1 based on the number of the current iteration.
Besides, we set $m=5$, beta distribution parameter $\beta = 0.75$ in mixup and $N=5, 10, 10, 50$ for \textbf{Office-31}, \textbf{Office-home}, \textbf{PointDA-10} and \textbf{VisDA}.
\emph{All results are the averages of three random runs with seed $\in$ \{0, 1, 2\}.}

\subsection{Comparison Results}
\noindent
\textbf{2D image datasets.} We first compare our method with the state-of-the-art methods on 2D image datasets in Table \ref{tab:home}, \ref{tab:office}, and \ref{tab:visda}.
Note that the results of other methods are from the original papers, except ours.
On \textbf{Office-home}, we achieve the best results on three tasks, and the highest mean accuracy, demonstrating the effectiveness of ProxyMix to deal with the multi-class classification problem on the medium-size dataset.
On \textbf{Office-31}, we also achieve the highest mean accuracy among SFDA methods, validating the efficacy of ProxyMix handling with small datasets with fewer categories.
On \textbf{VisDA}, we achieve the best results on four single tasks and a comparable mean accuracy with the state-of-the-art methods.
The reason why the performance on \textbf{VisDA} is not as good as the first two may be that the scale of the proxy source domain is too small relative to the entire dataset, which causes the network to have a certain bias towards the proxy source domain.
In summary, our method ProxyMix achieves competitive accuracy across three benchmarks when compared with others, which demonstrates the effectiveness in dealing with the standard 2D image domain adaptation benchmarks.
We achieve similar results compared with the state-of-the-art SFDA methods $\textrm{A}^2\textrm{Net}$ \cite{xia2021adaptive} (ICCV-21) and NRC \cite{yang2021exploiting} (NeurIPS-21), and UDA method ATDOC \cite{liang2021domain} (CVPR-21).
The presented results clearly demonstrate the efficacy of the proposed method in dealing with domain-imbalanced, multi-class and large-scale challenges.

\noindent
\textbf{3D point cloud dataset.}
To explore the generalization performance of ProxyMix on 3D data, we also report the results for the \textbf{PointDA-10} dataset in Table \ref{tab:pointda}. Without any extra modules, our method achieves the highest average accuracy on the benchmark, even compared with UDA methods and the 3D cloud point domain adaptation method PointDAN \cite{qin2019pointdan}.

\begin{table}
    \centering 	
    \caption{Analysis of different soft pseudo labels.}
    \resizebox{0.43\textwidth}{!}{     	 
    \begin{tabular}{lccccc}
    \toprule
    Choices of soft label & \textbf{Office-31} & \textbf{Office-home} &  \textbf{VisDA} \\
    \midrule
    MixMatch \cite{berthelot2019mixmatch} & 88.4 & 72.4 & 83.0 \\ 
    ReMixMatch \cite{berthelot2020remixmatch} & 88.1 & 71.3 & 80.2 \\
    ATDOC \cite{liang2021domain} & 88.5 & 72.2 & 84.7 \\
    Ours & \textbf{90.1} & \textbf{72.8} & \textbf{85.7} \\
    \bottomrule 
    \end{tabular} 
    }
    \label{tab:ab-fa} 
\end{table} 

\begin{table}
    \centering 	
    \caption{Analysis of aggregation strategy.} 	 
    \resizebox{0.43\textwidth}{!}{     	 
    \begin{tabular}{lccccc}     		 
    \toprule
    Variants & \textbf{Office-31} & \textbf{Office-home} &  \textbf{VisDA} \\
    \midrule
    w/o aggregation & 88.4 & 71.3 & 82.4 \\ 
    w/ aggregation (Ours) & \textbf{90.1} & \textbf{72.8} & \textbf{85.7} \\
    \bottomrule 
    \end{tabular} 
    } 	 
    \label{tab:ab-a} 
\end{table}

\begin{table}
    \centering 	
    \caption{Analysis of different selection methods of proxy source samples. } 
    \resizebox{0.4\textwidth}{!}{     	
    \begin{tabular}{lccc}     		
    \toprule     		
    Method & \textbf{Office-31} & \textbf{Office-home} &  \textbf{VisDA} \\     		
    \midrule     		
    Random-selected & 83.9 & 69.0  & 81.9  \\
    Entropy-guided & 86.3 & 70.5  & 72.6 \\
    Ours & \textbf{90.1}  & \textbf{72.8} & \textbf{85.7}   \\
    \bottomrule
    \end{tabular}
    } 	
    \label{tab:ab-s}
\end{table}

\begin{table}
    \centering 
    \caption{Ablation study on the loss functions.} 	
    \resizebox{0.4\textwidth}{!}{     	
    \begin{tabular}{cccccc}     		
    \toprule     		
    $\mathcal{L}_{ps}$ & $\mathcal{L}_{tgt}^{inter}$ & $\mathcal{L}_{tgt}^{intra}$ & \textbf{Office-31} & \textbf{Office-home} & \textbf{VisDA} \\ 
    \midrule     		
    $\checkmark$ & & & 83.5 & 66.3 & 69.6  \\ 
     & $\checkmark$ & & 89.1 & 72.4  & 78.5  \\ 
     & & $\checkmark$ & 86.7 & 65.8 & 84.9  \\  
    \midrule
    $\checkmark$ & $\checkmark$ & & 89.3 & 72.3 & 78.4  \\
    $\checkmark$ & & $\checkmark$ & 89.9 & 71.3 & 84.7  \\
    $\checkmark$ & $\checkmark$ & $\checkmark$ & \textbf{90.1} & \textbf{72.8} & \textbf{85.7}  \\
    \bottomrule
    \end{tabular}
    } 	
    \label{tab:ab-l}
\end{table}

\begin{figure} 	 
    \centering 	 
    \begin{tabular}{c}     		 
    \includegraphics[scale=0.7]{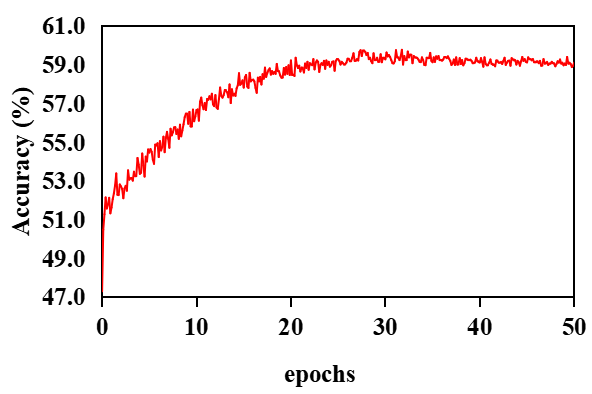}\\
    \end{tabular} 
    \caption{The accuracy curve of the task Ar$\to$Cl on \textbf{Office-home}.} 	 
    \label{tab:ab-acc} 
\end{figure}

\subsection{Empirical Analysis}
To explore the effectiveness of the proposed pseudo-labeling strategy, the aggregation strategy, the construction method of proxy source domain, we conduct a series of ablation analysis on the three common-used 2D image classification datasets \textbf{Office-31}, \textbf{Office-home} and \textbf{VisDA}. Then we explore the influence of three loss functions in our method, the training stability, and the sensitivity of the important hyper-parameters.
We also show the t-SNE visualization results of task Ar$\to$Cl to clearly validate the altering of features.

\noindent
\textbf{Effectiveness of the proposed frequency-weighted aggregation soft pseudo label.}
Our frequency-weighted aggregation strategy (FA) is a soft pseudo label generation method.
To verify the influence, we compare our method with three label refinery strategies.
1) MixMatch \cite{berthelot2019mixmatch} calculates the soft pseudo label by sharpening and normalizing the predictions directly.
2) ReMixMatch \cite{berthelot2020remixmatch} sharpens the predictions first, then multiplies a distribution alignment ratio calculated by the current batch of samples.
3) ATDOC \cite{liang2021domain} only uses the highest possibilities that are multiplied by a balanced ratio, causing the sums to not be equal to 1, which is not conducive to the calculation of KL divergence. Therefore, we normalize the predictions of ATDOC in our experiments.
The results shown in Table \ref{tab:ab-fa} demonstrate that the proposed frequency-weighted aggregation module effectively improves the soft label's reliability.

\begin{figure*}
    \centering   	 
    \begin{tabular}{ccc}     		 
    \includegraphics[scale=0.58]{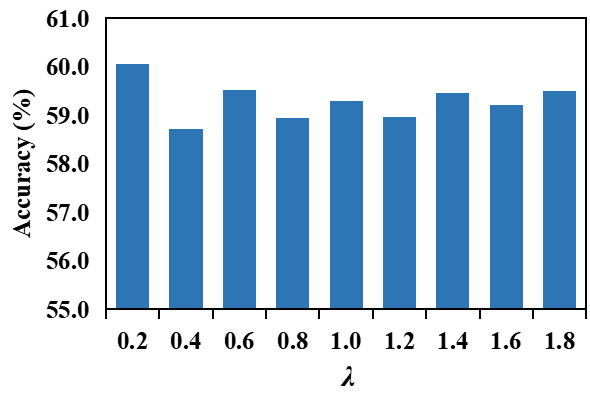} &
    \includegraphics[scale=0.58]{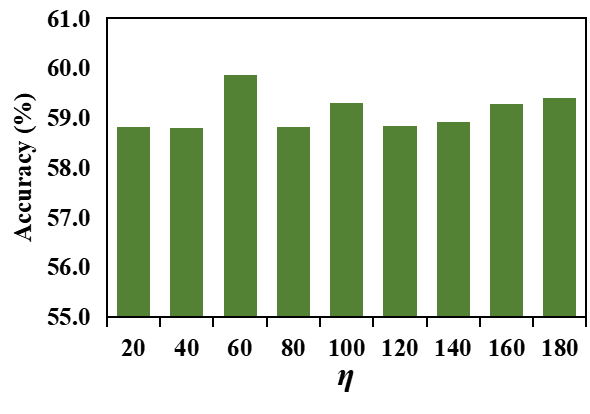} & 
    \includegraphics[scale=0.58]{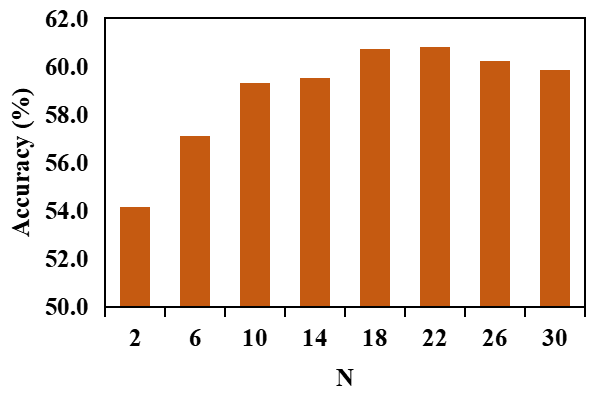} \\   		 
    (a) Influence of the weight $ \lambda$.  & (b) Influence of the weight $ \eta$.  & (c)  Influence of N. \\
    \end{tabular} 
    \caption{Sensitivity of hyper-parameters of task Ar$\to$Cl on \textbf{Office-home}. (a) Influence of the weight $ \lambda$ of $\mathcal{L}_{tgt}^{inter}$; (b) Influence of the weight $ \eta$ of $\mathcal{L}_{tgt}^{intra}$; (c) Influence of the number N per class in proxy source domain.}
    \label{tab:ab-fig} 
\end{figure*}

\begin{figure*}
    \centering
    \begin{tabular}{cc}
    \includegraphics[scale=0.35]{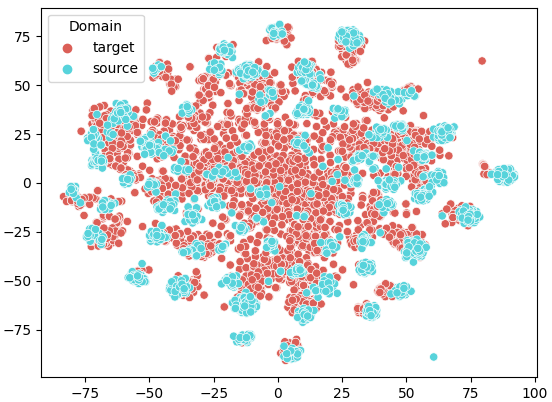} &
    \includegraphics[scale=0.35]{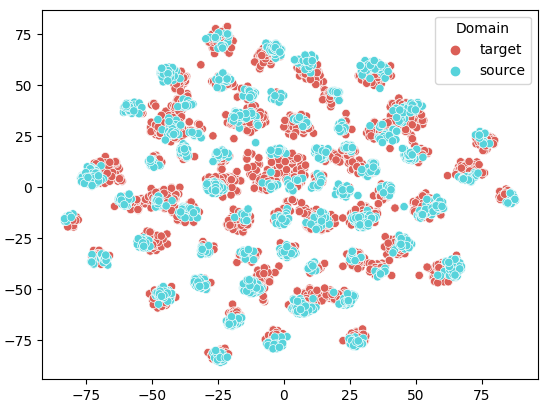} \\
     (a) Before adaptation. & (b) After adaptation. \\
    \includegraphics[scale=0.35]{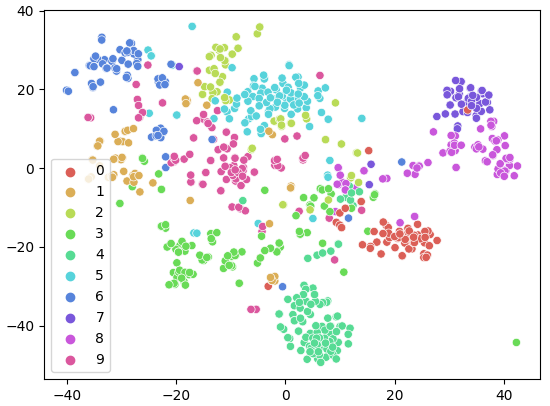} &
    \includegraphics[scale=0.35]{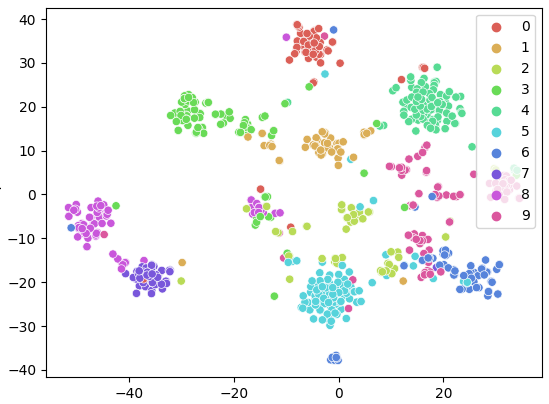}\\ 	 
     (c) Before adaptation. & (d) After adaptation. \\
    \end{tabular}
    \caption{The t-SNE visulization of task Ar$\to$Cl on \textbf{Office-home}. (a) and (b): the unseen source features (blue points) and the target features (red points) before and after adaptation, respectively. (c) and (d): the target features before and after adaptation, respectively. For clarity, we select first 10 classes in the 65 classes on \textbf{Office-home}.}
    \label{tab:ab-tsne}
\end{figure*}

\noindent
\textbf{Effectiveness of the aggregation strategy.}
Our aggregation technique pulls unlabeled target data to semantic neighbors, allowing us to investigate the target domain's structure information and mitigate the detrimental effects of noisy labels. 
Table \ref{tab:ab-a} shows the variant of ProxyMix without the aggregation approach to demonstrate the usefulness of the aggregation strategy. 
The accuracy of standard ProxyMix is higher than that of variants without aggregation, demonstrating that leveraging the semantic neighbors' center as the pseudo label is effective and reliable.

\noindent
\textbf{Analysis of the construction method of proxy source domain.}
To study the influence of the proposed construction method of the class-balanced proxy source domain, we compare ProxyMix with a common-used method, \textit{i.e.}, \textit{i.e.}, randomly-selected criterion, entropy-guided criterion, and the baseline method.
1) Randomly-selected: to ensure fairness, we randomly select N samples for each class from the target data to generate a class-balanced proxy source domain based on the classification results of the source model.
Because we cannot discover N examples for some difficult classes, we choose the remaining numbers of samples from other classes at random as compensation.
2) Entropy-guided: as commonly used in other works \cite{liang2021source}, we compare our method with the entropy-guided method.
In specific, we calculate the mean entropy $e$ of the source model's prediction on the full target dataset, then obtain a split ratio $\xi = \frac{ N(H(f_s(x_t))<e;x_t \in\mathcal{X}_t)}{N(x_t\in \mathcal{X}_t)}$, where $N(\phi)$ denotes the size of the subset formed by samples which satisfy the condition $\phi$, $H(\cdot)$ is the entropy function.
Then we compute the class distribution $\{n^{k}\}_{k=1}^K$ according to the predictions given by the source model, and select $n^{k}\cdot\xi$ samples with the lowest entropy for each class.
The results are shown in Table \ref{tab:ab-s}.
Random-selected perform unsatisfactory due to the poor confidence of the source model before adaption.
Although the entropy-criterion reflects the confidence of the prediction, it exacerbates the class imbalance problem and leads the model bias to the easier classes, which is not satisfactory in comparison to ours.
The proposed prototype-induced method achieves the highest accuracy. We take both confidence and class-balance into consideration, and as illustrated in Fig. \ref{criterion}, we observed that the accuracy of the proxy source domain is higher than the entropy-criterion.

\noindent
\textbf{Ablation studies on the proposed loss functions.}
To investigate the proposed loss functions, we show the results of variants with different combinations of loss functions in Table \ref{tab:ab-l}.
As shown, without the proxy source domain classification loss $\mathcal{L}_{ps}$, the accuracy of \textbf{Office-31} has the biggest drop.
The accuracy of \textbf{Office-home} is more likely to be influenced by the inter-domain mixup loss $\mathcal{L}_{tgt}^{inter}$.
As for the large-scale dataset \textbf{VisDA}, the intra-domain mixup loss $\mathcal{L}_{tgt}^{intra}$ contributes a lot.
The effectiveness of $\mathcal{L}_{tgt}^{inter}$ and $\mathcal{L}_{tgt}^{intra}$ also illustrate the reliability of the proposed frequency-weighted soft labels from another perspective.

\noindent
\textbf{Training stability.}
We show the accuracy curve of task Ar$\to$Cl on \textbf{Office-home} in Fig. \ref{tab:ab-acc}, the accuracy during training grows up quickly and then converges as we expected.
Therefore, the training procedure of ProxyMix is stable and reliable.

\noindent 
\textbf{Sensitivity of hyper-parameters.}
To better understand the effects of the hyper-parameters $\lambda$, $\eta$ and $N$, we explore their performance sensitivity in a single task Ar$\to$Cl on \textbf{Office-home} in Fig.~\ref{tab:ab-fig}.
The accuracies around $\lambda=1$ and $\eta=100$ fluctuate very softly in (a) and (b).
The results on the proxy source domain scale are provided in (c), shows that the accuracies change slightly around $N=20$.
Generally, in our method ProxyMix, the hyper-parameters are not sensitive.

\noindent
\textbf{t-SNE visualization.}
To evaluate the effectiveness of ProxyMix, We show the t-SNE visualization\footnote{\url{https://lvdmaaten.github.io/tsne/}} of target features on task Ar$\to$Cl in Fig.~\ref{tab:ab-tsne}.
To validate the effectiveness of domain alignment, we show the features of the unseen source domain (blue points) and the target domain (red points) in (a) and (b). The distribution of target features is closer to the source feature after adaptation as we expected.
We also show the target feature distribution of the first 10 classes of \textbf{Office-home} in (c) and (d). 
Benefiting from our frequency-weighted aggregation strategy, the feature clusters after adaptation are compact, and the classification boundary is clear.

\section{Conclusion}
In this paper, we focus on the source-free domain adaptation problem, and propose a simple yet effective method named Proxy-based Mixup training with label refinery (ProxyMix).
In specific, we treat weights of the fully-connected layer as class prototypes to choose a series of confident samples to construct a class-balanced proxy source domain.
Then label information is expected to flow from the pseudo source domain to the unlabeled target domain via mixup training.
To enhance mixup training, we further introduce a new pseudo label refinery strategy, which combines frequency-weighted sharpening and neighborhood aggregation to obtain reliable soft predictions of unlabeled target data.
Experiments on four popular benchmarks prove the effectiveness of ProxyMix without access to source data.
Although our method outperforms several UDA methods that are based on source data, we should recognize that removing all noisy labels in an unsupervised manner is still tough. 
We believe that our work is an attempt in that direction, with the intention of inspiring others in the UDA community.

\bibliographystyle{IEEEtran}
\bibliography{IEEE}

\end{document}